# Generating machine-executable plans from end-user's natural-language instructions

Rui Liu, Xiaoli Zhang*

*Department of Mechanical Engineering, Colorado School of Mines, Golden, Colorado, 80401, USA*
*{rliu, xlzhang}@mines.edu*

*Abstract*- It is critical for advanced manufacturing machines to autonomously execute a task by following an end-user's natural language (NL) instructions. However, NL instructions are usually ambiguous and abstract so that the machines may misunderstand and incorrectly execute the task. To address this NL-based human-machine communication problem and enable the machines to appropriately execute tasks by following the end-user's NL instructions, we developed a Machine-Executable-Plan-Generation (exePlan) method. The exePlan method conducts task-centered semantic analysis to extract task-related information from ambiguous NL instructions. In addition, the method specifies machine execution parameters to generate a machine-executable plan by interpreting abstract NL instructions. To evaluate the exePlan method, an industrial robot Baxter was instructed by NL to perform three types of industrial tasks {"drill a hole", "clean a spot", "install a screw"}. The experiment results proved that the exePlan method was effective in generating machine-executable plans from the end-user's NL instructions. Such a method has the promise to endow a machine with the ability of NL-instructed task execution.

*Index Terms*-Advanced manufacturing machine, machine-executable plan, natural language instruction, semantic analysis, task execution.

## 1. Introduction

Human-machine collaborative manufacturing combines human intelligence on high-level task planning and the robot physical capability (e.g., precision and speed) on low-level task execution [1]. Toward this direction, intuitive and natural communication between the human and the machine has been an active research area in the last decade with the goal to enable seamless human-machine cooperation [2][3]. Natural-Language-instructed human-machine interaction is expected to enable an advanced manufacturing machine, such as a Computer Numerical Control machine or an industrial robot, to autonomously perform tasks such as rough/fine finishing [4][5], assembly [2][6] and packaging [7][8] according to the end-user's NL instructions, which are given based on the user's judgement of the task progress and environmental situations. Compared with other input methods, including human hand force [9][10], hand gesture [11][12], and body motions [13][14][15][16], the NL instruction method has two main advantages. First, NL instruction provides a natural, human-like, face-to-face communication manner. Non-expert users without prior programming training could command a machine to perform their desired tasks [17][18]. Second, the inherent linguistic structure of NL, as a predefined information encoder, provides a standard, informative data source to generate structured machine language [19][20]. In contrast, the aforementioned existing methods require extra translations among discrepant data patterns. These two advantages make NL a superior means for the end-user to naturally and efficiently communicate with manufacturing machines.

Currently, typical industrial applications involving NL include NL-based control in which the working statuses such as "on/off" and "quickly/slowly" are selected orally to control a machine in navigation [3][21], NL-based task execution in which the task operation methods such as "goTo + Location; then drop + object" is described orally to help a machine with object finding/placing [22][23], and NL-based execution personalization in which human's preferences and moods in oral dialogs were considered to adjust a machine's execution manners [24][25].

However, there is still a long way to apply NL-instructed machines in practical manufacturing applications. First, NL is variable and ambiguous. NL is usually polysemous, homophonic and expression-manner diverse so that the same meaning could be expressed in various ways, and different meanings could be expressed in similar ways [8][26]. For example, "drill a hole" could be expressed as "bore one hole", "drilling one bore", "create an unthreaded hole", and so on [27]. In addition, humans usually use referring, outlining, and omitting in NL instructions [22][28]. For example, in an instruction "at the center point", information such as "which object in which place has the center point" cannot be known merely from a word 'the' [27]. It is challenging to extract task-related information such as task goals and detailed execution procedures from NL instructions. Second, human instruction is abstract [23][29]. Even when a complete execution procedure for a task is instructed, the generated plan is still non-executable for a machine. For example, the abstract instructions 'clean the surface' are still machine-non-executable for that the execution-related specific knowledge such as "tool: brush; action: moveDown → sweep → moveUp; ..." is missing [27]. In addition to specific-knowledge missing, a reasonable and flexible knowledge structure, which is implicitly embedded in NL descriptions to guide correct task execution, is difficult to extract [30][31][32]. By obeying the human instructions, one task could be flexibly executed by several methods, which were formulated according to an individual's cognitive logics [33][34]. However, usually these cognitive logics in NL instructions are difficult to understand as to a machine, for that literal information directly extracted from NL instructions is insufficient to explain the logics [2][35]. Taking the task "deliver a drink" as an example, the potential methods could either be "fetch a cup + fill the cup with water + place it on table" or "place a cup on the table + fill cup with water". The logics such as {*CupAvailability(yes)* ∧ *WaterAvailability(yes)* → *CupBeingFilledFeasibility(yes)*} behind the task executions have not been described explicitly in NL instruction while these logics are important in deciding what kind of procedures are feasible and reasonable in execution and in assessing whether a task could be executable or not. It is challenging for a machine to perform a task without knowing the task-related logics.

---

*Corresponding author: Tel.:+1-303-384-2343; fax:+1-303-273-3602, email:xlzhang@mines.edu.

To address these problems and enable NL-instructed manufacturing in practical industrial tasks, we developed a machine-executable-plan-generation (exePlan) method to "translate" the ambiguous and abstract NL instructions into machine-executable plans. In this paper, we mainly have two contributions, shown as follows.

• A task-centered semantic analysis method is developed for processing ambiguous NL instructions into task-related information including task goal, sub-goals, and execution logic relations. Instead of using basic linguistic features such as keywords/Part-of-Speech(PoS), the task-related semantic features, such as actions/tools/execution logics were considered to extract the task-related information from ambiguous NL instructions.

• A machine-execution-specification method is developed for interpreting abstract human instructions into machine-executable plans. With this method, each abstract sub-goal in the NL instruction is firstly specified into an executable sub-goal by adding the machine-execution specification (MES) parameters such as location, action, and human requirements. Then a machine-executable plan is specified by exploring the weighted logic relations among the task-related execution procedures. Different from the first-order logic in which all the logic relations are inviolable and plans using first-order logic have binary executability {executable, non-executable}, weighted logic relations could be violable with a corresponding weighted decrease of plan executability and the plans using weighted logics have a range of acceptable executability. A plan is flexibly made by organizing reasonable logic procedures with the executability greater than a threshold value.

## 2. Related Work

To disambiguate NL instructions in task execution, special grammars were designed to identify the task-related entities based on specific keyword involvements and their PoS tags. For example, in the sentence "bring the can in the trash bin" the task goal "in the trash bin" was extracted based on the keywords "bring, can" and their corresponding PoS tags "VB, NN'[8]. Ontology relations among the interested entities were used for mutual disambiguation. For example, to describe a cup, the description was likely to be "container with handle attached". "Attached" was the constraint relation between the object "container" and object part "handle" [36][37]. When an entity was ambiguously mentioned, the ambiguous entity could be explained by mutually co-referring. Take sentences "Go to the second crate on the right. Pick it up" for an example, with co-reference resolution the uncertain expression "it" was identified as "the second crate on the right" [22][38]. When the NL descriptions such as "pick up the pallet" were too ambiguous for a robot, a query such as "which pallet?" was launched to ask the human for disambiguation [2][39]. By exploring the features such as perceivable properties "cylindrical" and "round", the ambiguous descriptions "cylindrical container with a round handle attached on one side" for the object "container" was understood [36]. By exploring the spatial relations "behind" in NL descriptions "Navigate to the building behind the pole", named entities "building, pole" were identified in the real world [40][41]. To disambiguate the NL instructions, these methods explored context evidences for a single entity. Evidences include semantic relations, human explanations, and spatial constrains. However, these methods only focused on using one single type of evidences such as basic linguistic feature keywords/PoS or semantic features co-referring and perceivable properties, without combining the multiple types of features together to perform a comprehensive semantic analysis. In addition, these methods aimed to identify an entity such as "can" or "trash bin" separately without considering entity correlations such as "can—trash bin", which are informative in instruction disambiguation. The above mentioned features are important for semantic analysis, however have not been well investigated.

To interpret abstract expressions in NL instructions, motion grammars were first designed for establishing the word-action correlations such as word "grasp" — action "Grasp" [3][21][23]. Real-world preconditions such as "stay in the kitchen" were defined for triggering specific types of executions such as "visiting the kitchen" [17]. The NL descriptions were marked by landmark objects such as "staircase, box" in the real world to enable the execution of tasks such as "reach in a spot" [42][43]. With these methods, abstract NL descriptions are interpreted into executable commands to some extent. However these methods do not make the NL command truly machine-executable for that the critical execution parameters, including the tool usage, real-world precondition, action sequence, and human requirements, are still missing or insufficient for supporting a robot's executions in practical situations.

To interpret implicit cognitive logics embedded in NL instructions, probabilistic graphical models were designed to explore the knowledge importance with the consideration of its probability distributions for plan making. For example, in the NL descriptions "go to the second crate on the right. Pick it up.", the procedures could be modeled as {goTo create (p=0.50), PickUp crate (p=0.50)}[22][43]. In [46][47], a semantic topological model was developed to explore the internal logic correlations of sub-steps in a reasonable task-execution plan. For example, in the path "first go to the hallway; the cafeteria is down the hallway" the hallway could be replaced by "hall, corridor, walkway" and the "cafeteria" could be replaced by "dining hall" according to the semantic topology. The procedures with any combinations of the elements in the topology structure were considered as reasonable task execution plans. However, these plans are not truly executable for a machine. Probabilistic graphical models merely describe the importance of the sub-steps in a plan, ignoring their internal logics without which a plan is non-executable in the real world. On the other hand, topological models describe the logic relations among procedures; however, the logic constrains are hard without discriminative descriptions of the involved logic relations. If one hard logic relation is unsatisfied the whole plan is not executable. Both probability-based methods and logic-based methods should be considered to solve the plan-making problem. However, they have not been well investigated.

## 3. exePlan Methodology

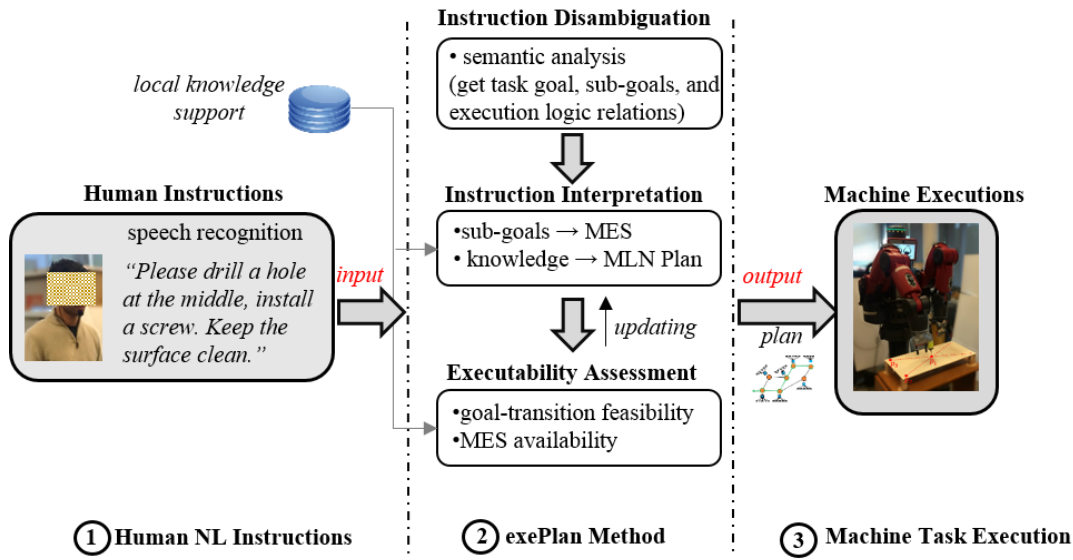

Fig. 1. Framework for the exePlan method. With the exePlan method, NL instructions are translated into machine executable plans. MES denotes machine-executable specifications, and MLN denotes the markov logic network representation.

As Fig. 1 shows, the exePlan method transfers NL instructions (input) into machine task execution plans (output). The exePlan method includes three critical steps: instruction disambiguation, instruction interpretation, and plan executability assessment. Instruction disambiguation conducts task-centered semantic analysis on NL instructions to extract task-related information including task goal, sub-goals, and execution logic relations. Instruction interpretation generates machine-executable plans from abstract task-related information by specifying MESs for each sub-goal and organizing the sub-goals into a logic-based plan. The main types of MES parameters include working-spot locations, working-spot statuses, action sequences, tool usages, and operation preconditions. A logic-based plan represents a reasonable execution procedure in a weighted logic manner. Executability assessment checks feasibility of generated sub-goals, sub-goal transitions, and the MES availability in the practical situations, enabling the plan to be machine-executable.

3.1. Instruction Disambiguation

Initial natural language processing (NLP) is conducted for processing NL instructions into hierarchical syntax trees with linguistic features such as words, PoS tags, word dependencies, and independent sentences. First, English oral instructions are recognized as text corpus by the speech recognition tool SpeechRecognition [48]. Second, text corpus is split into independent sentences, words, PoS tags and dependences by using the NLP tool Stanford CoreNLP [20]. Then word normalization is conducted to normalize the interested keywords such as "drilling, drills and drill" to unified morphologies such as 'drill'. Unified formats of the linguistic features are for preparation of feature-based semantic analysis. Last, based on the linguistic features, sentence structures are analyzed by generating the hierarchical syntax tree, shown in Fig. 2 with {root: sentence root, dobj: direct object dependency, nomd: norm modifier dependency, amod: adjectival modifier, case: case dependency, det: determiner dependency, VB: verb, NN: noun, DT: determiner, JJ: adjective, IN: preposition}.

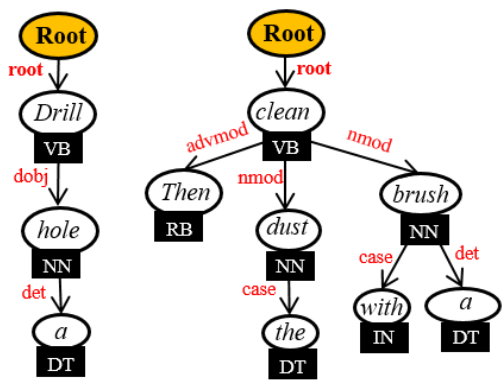

Fig. 2. Syntax analysis for the sentence "Drill a hole. Then clean the dust with a brush." Sentence structure is analyzed based on the linguistic features including sentence root, words, PoS tags and word dependencies.

Task-centered semantic analysis aims to identify task-related entities such as execution sub-goals and sub-goals' logic relations indicated by ambiguous NL instructions. A task-centered feature space is designed by using both the basic linguistic features {keyword (w), PoS (pos), word dependencies (wd), general words' PoS dependencies (posd)} and the task-related semantic features {task-related keyword (KW), previous/next sub-goal (preSubg/nextSubg), previous/next noun of a task-related entity (preN/nextN), previous/next word of a task-related entity (preW/nextW), dependency of the task-related keywords (KWD), task-related keywords' PoS dependency (posD), dependency of the task-related norn (NND), and task-related context (context)}. By projecting the NL instructions into this task-centered feature space and performing classifications, sub-goals in an instructed execution plan are identified. Then based on the identified sub-goals, logic transition relations among the sub-goals are detected according to their involvement sequence. Taking the instruction "*Drill a hole at the board center. Then clean the dust with the brush.*" as an example, the sub-goal "clean" is identified by using features {KW: clean, pos: VB, KWD: clean+dust, posD: dobj(VB+NN), preSubg:(drill), context: (dust, brush)}. Then based on the sub-goals' temporal sequence in human instructions, the logic transition "drill→clean" is detected, shown in Fig. 3.

Keywords for the task-related entities are defined in the local database. When keywords are detected in NL instructions, semantic analysis is triggered for knowledge extraction. Given that the amount of the text corpus is potentially large and the desired output predictions (sub-goals involved in task plans) are multiple, a semi-supervised multi-class Support Vector Machine algorithm (smSVM) [44] is adopted. A semi-supervised classification method merely needs a small amount of labeled samples (usually 1%~5% of the total samples) for classifier training [32]. During the semi-supervised training, new features are learned for better entity-extraction performances. The detailed process of the smSVM-supported entity detection is shown in Algorithm 1. First, by using the manually labeled sentence vectors $(P_k, Q_k)$, a classifier h for an entity is initially trained. $(P_k, Q_k)$ is represented by $(p_{1:k}, q_{1:k})$, where $p_k$ denotes the k-th feature vector of a sentence in human instructions and $q_k$ denotes the corresponding entity label. Second, the initially-trained classifier $h$ is adopted to classify testing samples $q_{k+1:m}$ into label $p_{k+1:m}$ one by one.

After a classification for a single sentence, the classified sample ($p_{k+1}$, $q_{k+1}$) is used to update the existing training sample set into a new set T, based on which the classifier *h* is updated. In the new training process, new semantic features for a type of classification are collected to improve the classifier's performance. The core algorithm for smSVM is shown in equations (1) and (2). $w_{sm}$ is the weight for an input feature vector, $\xi_i$ denotes the acceptable classification error, and $C_{sm}$ is the tradeoff parameter balancing error and margin. The slop value $w_{sm}^*$ and the intercept value $b_{sm}^*$ are solved by a quadratic programming (QP) solver [45] and they define an optimal hyperplane to conduct classification for extracting the task-related entity $f_{sm}$.

$$\min \tfrac{1}{2}||w_{sm}||^2 + C_{sm} \sum_{i=1}^{K} \xi_i \qquad (1)$$

$$f_{sm}(x) = sign\{w_{sm}^{*T} \cdot x + b_{sm}^*\} \qquad (2)$$

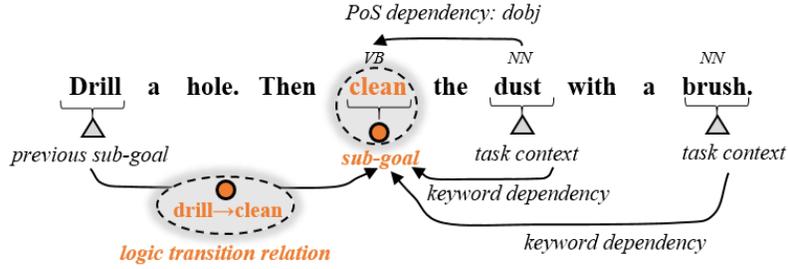

Fig. 3. Task-centered entity detection for identifying sub-goal "clean a hole"

---

**Algorithm I Semi-supervised Task-related Entity Detection**

Abbreviations: sentence vector *p*, sentence label *q*, sentence vector set *P*, entity label set Q, entity feature function φ
**Input:** labeled sentences $(P_k, Q_k) = \{(p_{1:k}, q_{1:k})\}$, unlabeled sentences $P_m = \{p_{k+1:m}\}$, testing sentences $P_j = \{p_{j:}\}$
**Output**: classifier *h*, labeled sentence set *T* and precision

Let training set $T=(P_k, Q_k)$, classifier $h_0$ with SVM, where **arg max** $h_0(\varphi(P_k)) = Q_k$
classifier $h=h_0$
**for** $p_l \in P_m$ **do**
    features $z_l = \varphi(p_l)$
    new label $q_i$ = **arg max** $h(z_l)$
    new training set $T \leftarrow T + (p_l, q_l)$
    Training new classifier h: **arg max** $h(\varphi(T)) = Q_k$
**end for**
precision = proportion of correctly predicted labels in total predicted labels
**return** *h*, *T*, precision

---

3.2. Instruction Interpretation
1). Sub-goal Interpretation

An abstract sub-goal is interpreted into an executable operation by specifying the corresponding MES parameters. A complete MES consists of operation preconditions (precon), working spot locations (loc), action sequence (act), tool usages (tool), and human requirements (req), shown in Fig. 4. The MES parameters are hard constraints for a sub-goal that only when all the hard constraints are satisfied, a sub-goal is executable with interpretation value 1; otherwise the interpretation value is 0. As shown in equation (3), $mes_i$ denotes a MES parameter for sub-goal *i* of task *j*, n is the total number of MES parameters for sub-goal *i*, and $subgoal_i$ denotes the binary executability for the *i*-th sub-goal. In the practical implementation process, NL instructions specify some MES parameters, and the interpretation module recommends the remaining MES parameters.

$$\forall i,j \ [ \bigwedge_{i=1}^{n} \textbf{mes}_i \rightarrow \textbf{subgoal}_i(task\ j) \ ] \tag{3}$$

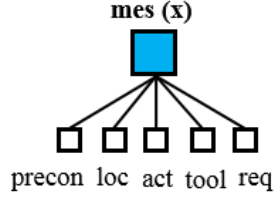

Fig.4. Sub-goal interpretation by machine execution specifications (MES)

(2) Execution Planning

An execution workflow is the machine-executable task procedure consisting of each executable sub-goal generated from the previous step. The generation of this workflow relies on the task-centered cognitive logics that are learned from NL instructions. These logics are the key to enable the exploration of the operation correlations among the task sub-procedures given specific environmental conditions and human preferences. The task-centered logic framework is modeled by a Markov Logic Network (MLN) algorithm [49][50] (shown in equation (4)). In this algorithm, each sub-goal and sub-goal transition are expressed as logic clauses $f_i$ with weighted contribution $w_i$ towards task execution, $n_i$ denotes the total number of clauses $f_i$ which are satisfied by the real-world condition $x$, the clause set is denoted by $\mathcal{F}$, and z is the total number of all the possible worlds of grounding the logic clauses with the real-world condition $x$. Different from hard constraints in first-order logic, soft constraints in a MLN describe the contribution of logic relations in a weighted manner, making the task execution flexible. This means, if some human-instructed logic transitions are inconsistent with the pre-trained logic transitions, the task is still likely to be executable with a lower value of executability, instead of being non-executable., including all logic clauses. A potential executable plan is generated by combining the grounded logic clauses according to human instructions. The optimal execution plan is the logic flow with the maximum weight accumulation. A sample MLN logic flow for the task "drill→clean" is shown in equation (5). The corresponding execution feasibility is calculated by equation (6), where the numbers denote the corresponding contributions of the logic formulas "Drill, TransitionFeasible, Clean" towards the task "drill→clean". Task "drill→clean" is executable only when its significant logics "drill is executed (Drill), clean is executed (Clean), and transition among drill and clean is feasible (TransitionFeasible)" are satisfied, the task is executable with the executability 0.9.

$$P(X = x) = \frac{1}{Z} \exp\left(\sum_{f_i \in \mathcal{F}} w_i n_i(x)\right) \tag{4}$$

$$Drill(mes_1) \wedge TransitionFeasible(mes_1, mes_2) \wedge Clean(mes_2) \Rightarrow Task(mes_1, mes_2) \tag{5}$$

$$0.3+0.3+0.3 \Rightarrow 0.9 \tag{6}$$

The most likely execution plan could be made by mapping the human-instructed procedure into MLN structures for different tasks and selecting the most reasonable procedure with the greatest execution feasibility. Execution planning is formally expressed by equation (7). As Fig. 5 shows, the green plan is the mapped human-described plan for a specific task.

$$\begin{aligned} y_{max} &= \arg\max_{y} \frac{1}{Z} \exp\left(\sum_{f_i \in \mathcal{F}} w_i n_i(x,y)\right) \\ &= \arg\max_{y} \sum_{f_i \in \mathcal{F}} w_i n_i(x,y) \end{aligned} \tag{7}$$

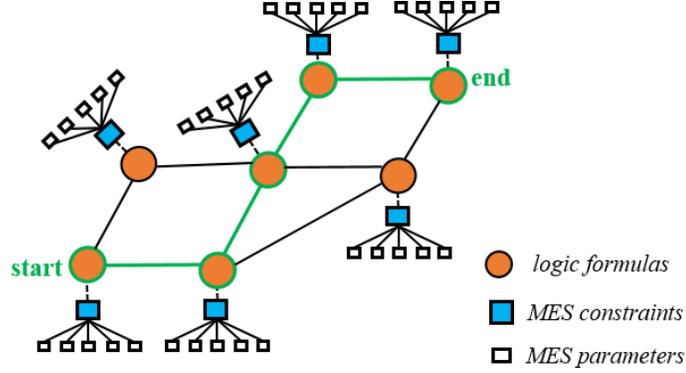

Fig. 5. MLN-based execution plan representation. Each circle denotes a logic clause (or execution sub-goal). Lines among circles denote logic relations among sub-goals. Green line denotes the machine-selected optimal execution plan, where "start" denotes the first execution step and "end" denotes the final execution step. With a MLN, execution steps' logic relations and their contributions towards task executability are described.

Learning the MLN representation for a task's execution includes two steps. The first step is defining the logic formulas for a task execution by domain experts. Detailed formula definitions are shown in section 4.1 task knowledge collection, Table 1. The second step is adopting a Structural Support Vector Machine (SSVM) method [50] to learn the formula involvements and their corresponding weights given a task. SSVM is good at learning complex structures, which are constructed by knowledge entity involvements and their weights [51][52]. In this paper given each task, each sub-goal only needs to be involved once to meet the basic execution requirements of a task. $n_i$ in equation (7) is represented as feature function $\Psi(x,y)$ which means when a logic clause is true the feature function value is 1, otherwise 0. With real-world constraints, the weight learning problem of the MLN is formulated as a Quadratic Programming (QP) problem in which the optimal weights are solved to achieve the maximum mutual task differences. A SSVM aims to learn two important aspects of the MLN representation, logic formula involvements and their weights. Logic formulas for sub-goal transitions are defined by human experts and extracted by the task-centered semantic analysis. In MLN learning, our objective is to learn the optimal MLN structure S: $\mathcal{X} \rightarrow \mathcal{Y}$ (training samples $(x, y) \in (\mathcal{X} \times \mathcal{Y})^n$), based on which the most executable plan $\mathcal{Y}$ (output) is predicted from NL-instructed task knowledge $\mathcal{X}$ (input) with the smallest error.

Given that a plan's executability is accumulated from the products of feature involvements and their weights, equation (4) could be simplified into a discriminant function $f_w$, shown in equation (8) where $W$ is a weight matrix $\mathcal{R}^N$ and $\Psi$ is the feature function expressed by $x$ and $y$. Then the MLN learning aims to find the weights W for minimizing the prediction error of equation (7). This weight learning problem could be formulated as a 1-slack SSVM problem and solved by an efficient cutting plane method [51], formulated by the objective function in equation (9), where $C$ denotes the trade-off parameters between the error $\xi$ and the margin $W$. Equation (10) denotes the constraint for the cutting-plane optimization problem to ensure that the discriminative value of the correct prediction $y_i$ is greater than the discriminative value of the incorrect predictions $\bar{y}_i$. To simplify the learning process, the calculation is only conducted between the correctly-predicted task $y_i$ and the most-executable incorrectly-predicted task $\bar{y}_i$, defined by equation (11). $\Delta(y_i, \bar{y}_i)$ is the loss function, which is defined as the executability difference between tasks $y_i$ and $\bar{y}_i$.

$$f_w(x,y) = W^T \Psi(x,y) \tag{8}$$

$$\min_{W, \xi \geq 0} \frac{1}{2} W^T W + C \xi \tag{9}$$

$$\text{s.t. } \forall (\bar{y}_1, \bar{y}_2, \ldots, \bar{y}_n) \in \mathcal{Y}^n, \frac{1}{n} W^T \sum_{i=1}^n (\Psi(x_i, y_i) - \Psi(x_i, \bar{y}_i)) \geq \frac{1}{n} \sum_{i=1}^n \Delta(y_i, \bar{y}_i) - \xi \tag{10}$$

$$\bar{y}_i = \operatorname*{argmax}_{y \in \{y_1, y_2, \ldots\}} w^T \Psi(x_i, y) + \Delta((y_i, y)) \tag{11}$$

3.3. Executability Assessment

To make the human-instructed plan executable in practical situations, executability for sub-goals, sub-goal transitions and MES availabilities were assessed. If the MES is incomplete or the overall execution likelihood is lower than a threshold $\partial_0$, the task is non-executable and the instruction interpretation process is triggered to make the involved sub-goals executable by filling in the missing MES parameters; otherwise the task will be executed. After interpretation, if the executability is still lower than the threshold $\partial_0$, a human instructor will be notified of the execution failure.

## 4. Evaluation

To evaluate exePlan's effectiveness in generating machine-executable plans, an advanced machine, the humanoid robot Baxter, was instructed to execute three types of tasks {drill, clean, install} at three different locations {center spot, upper-right spot, bottom-right spot} in a lab environment. The experiment platform was set up as shown in Fig. 6, where a human instructor orally commanded a robot to execute tasks on 3 targeted points on a working surface. The available tools included a microphone, a Kinect sensor, a brush, a screwdriver, and a driller. The detailed human-robot interaction process is described in the caption of Fig. 6. The three types of tasks {drill, clean, install} were assigned by the instructor based on his execution and expression habits. Depending on different task situations and different instructors, the NL instructions could vary, such as {clean a spot, drill a hole, install a screw, drill a hole → clean the dust, install a screw → clean the dust, drill a hole → install a screw, drill a hole → install a screw → clean the dust, drill a hole → clean the dust → install a screw, …}. Therefore due to their variability, these three types of tasks were selected to evaluate the effectiveness of the exePlan to generate machine-executable plans in NL-based human-machine interaction.

With the above-mentioned experiment setup, we aimed to evaluate two main aspects of our exePlan method. First, the accuracy of task-related information extraction evaluated the effectiveness of instruction disambiguation. Second, the plan identification accuracy and plan executability evaluated the effectiveness of instruction interpretation.

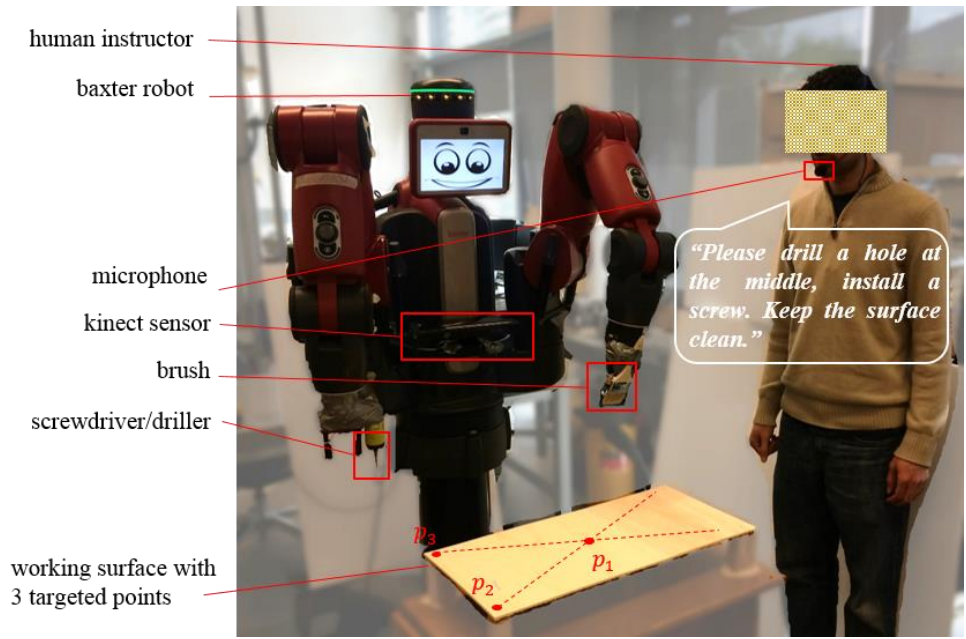

Fig. 6. Experiment platform setup. With a microphone, the human instructor gave NL instructions to assign a task for the advanced machine, a Baxter robot. The location and surface status of the targeted working spot were identified by a Kinect sensor. By speech recognition conducted on a local computer, oral instructions were translated into text corpus, based on which the robot analyzed human instructions to understand the assigned task. Then with the semantic meaning analysis of the instruction as well as the real-world conditions, human instructions were interpreted into machine-executable plans.

### 4.1. Task Knowledge Collection

To collect the MLN-based task knowledge for machines' executions, 800 volunteers were recruited on the crowdsourcing website Amazon Mechanical Turk (AMT) [53] to give 2400 copies of NL instructions (about 4100 task-related sentences) for executing the three types of tasks mentioned in prior experiment setup. Two types of questions were asked in the instruction collection process. One type of questions was at the overall plan level, such as "For instructing a robot to perform the task 'install a screw at the upper-right spot', please give your step–by-step instructions". The complete execution plan and some of the important execution parameters such as working spot location, necessary tools, and actions were provided by the volunteers. The second type of questions was at the MES level, such as "please describe the five types of MES parameters {location, actions sequence, tool usage, human requirement, working precondition} according to your expression habits". In this question, different MES parameters and their various expression manners for the three tasks were collected. With these instructions, the MLN structure and the related MES parameters were learned to represent the targeted tasks. In both MES collection and implementation, MES parameters were detected by the related keywords, such as "screwdriver, screw drive, screwdrivers, …" for the tool "screwdriver" and "clean, cleans, sweep, sweeping, remove, …" for the action "clean".

The likely logic formulas for MLN representations of the three types of tasks were defined by two expert volunteers, shown in Table 1. With these logics, a task was decomposed into small sub-goals. To finish a task, some of these goals must be achieved by satisfying the corresponding logics. The weights of these predefined clauses were learned by a SSVM to construct the MLN representations. These logic formulas consist of three basic execution steps {DrillHole, CleanSpot, InstallScrew} and their mutual logic transitions. Given that one task could be executed by multiple execution manners, MLN learning aims to learn the relative importance of the logic formulas given a task. The benefits of learning formulas' relative importance is that some important formulas could be flexibly involved with the consideration of both their importance and internal logic relations in task planning, increasing task planning flexibility and accuracy.

Table 1. Logic formulas and their definitions

| Formula | Definition |
| --- | --- |
| CleanSpot | sub-goal clean is feasible |
| DrillHole | sub-goal drill is feasible |
| InstallScrew | sub-goal install is feasible |
| tranCD | sub-goal transition from clean to drill is feasible |
| tranDC | sub-goal transition from drill to clean is feasible |
| tranCI | sub-goal transition from clean to install is feasible |
| tranIC | sub-goal transition from install to clean is feasible |
| tranDI | sub-goal transition from drill to install is feasible |
| tranID | sub-goal transition from install to drill is feasible |

In the semi-supervised entity detection, classifiers for three basic formulas {*DrillHole*, *CleanSpot*, *InstallScrew*} were trained. To filter out the irrelevant instructions, a fourth type of basic formula "Other" was added to denote the instructions that donot include the three basic logic formulas. A set of 50 sentence samples are manually labeled as seed samples for initially training the classifiers. A set of 4000 unlabeled sentence samples was prepared for further training the classifiers. Another new set of 50 samples was manually labeled for evaluating the classifiers' real-time performances during the semi-supervised learning process. After training, the top 20 influential semantic features for the classifiers of the four formulas are shown in Fig. 7. The influential features such as keywords "drill", keyword dependency "make+hole" and next sub-goal "clean" given a basic formulas such as "DrillHole" are reasonable according to our NL expression habits.

Based on the trained classifiers of the three basic formulas {DrillHole, CleanSpot, InstallScrew} as well as the 2400 copies of NL instructions, the weights of the defined formulas given each task were learned. The MLN representation for the three tasks is shown in Fig. 8. The corresponding formula features and their weights for different tasks are shown in Fig. 9. Taking the task type "install" as an example, the relative important logic formulas are "InstallScrew" and "transCI" which denote that "screw" and "cleaning the surface" are relatively important in finishing the "install" task. This result is consistent with our daily common sense. The collected MES parameters for the three basic formulas are shown in Table 2.

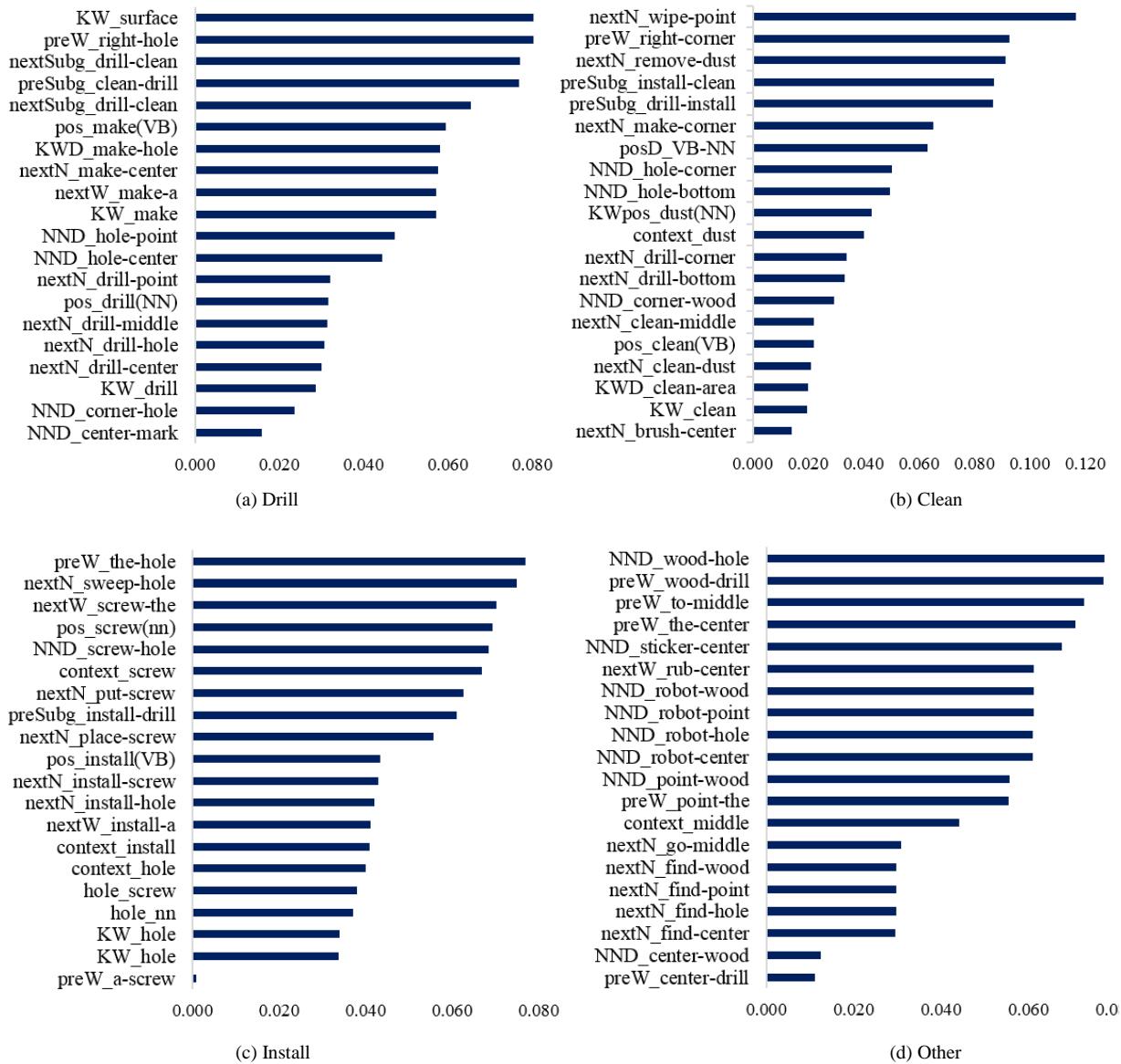

Fig. 7. The top 20 influential semantic features of the entity-detection classifiers

*4.2. Evaluation of Intruction Disambiguation*

A new set of 20 volunteers were recruited to instruct the Baxter robot in the lab environment (shown in Fig. 6). Each of the volunteer was required to naturally describe three tasks for the robot to execute, shown in Fig. 10. Supported by the exePlan method, the Baxter robot was able to understand task-related logics from the ambiguous human instructions. To evaluate the performances of the task-centered semantic analysis by using the exePlan method, an algorithm-level baseline was selected as Naïve Bayesian (NB) [55] which is efficient and classic in performing classification and a method-level baseline was selected as nonTC [8][54] which only considers the keyword-related features in semantic understanding. Accuracy of task-related logic formula detection was used to evaluate exePlan's effectiveness in understanding a task from the ambiguous NL instructions. Accuracy was assessed by precision and recall. Recall was calculated by the percentage of the machine-extracted entities in the overall involved entities. Precision was calculated by the percentage of correctly identified entities in the overall identified entities. As Table 3 shows, TC's performance was good in both the SVM-based method and NB-based method, with recall higher than 0.97 and precision higher than 0.96. The TC method consistently outperformed the nonTC method in both recall

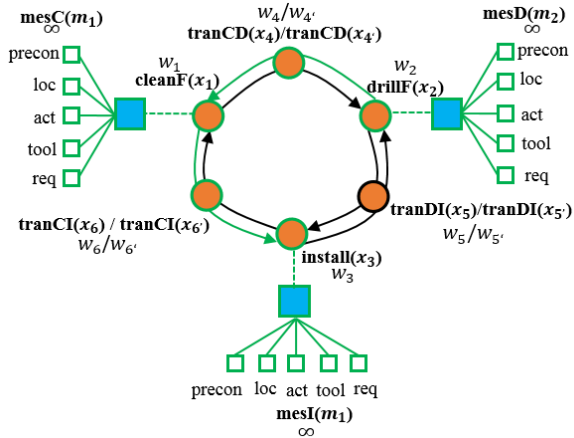 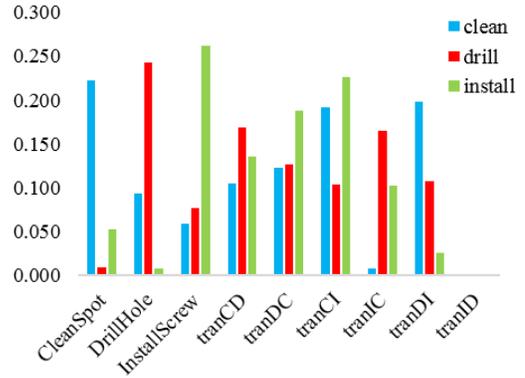

Fig. 8. MLN-based execution workflow representation. Each circle denotes a logic clause. Lines among circles denote logic relations among sub-goals. Green line denotes the machine-selected optimal execution plan. Each sub-goal has a weight, serving as a soft constraint. Each MES parameter has an infinite weight, serving as a hard constraint.

Fig. 9. MLN formula weights for the three types of tasks "clean, drill, install". For a type of task such as "drill", different logic formulas {CleanSpot, DrillHole, InstallScrew, TranCD, TranDC, TranCI, TranDI, TranID} contribute to the task executability with different weights {0.009, 0.243, 0.077, 0.169, 0.126, 0.104, 0.165, 0.108, 0.002}. The formulas "drill a hole (DrillHole), transition from 'clean' to 'drill' (transCD)" are relatively important for a "drill" task.

Table 2. Machine Execution Specification (MES) details

| Formulas | MES Types | Detailed MES parameters |
| --- | --- | --- |
| CleanSpot | precondition (precon) | spot dirty; after drilling |
| | location (loc) | upper-right; center; bottom-right |
| | action sequence (act) | place→sweep/clean/rub→left |
| | tool usages (tool) | brush; air hose; cloth; rag; broom; sweeper |
| | human requirements (req) | sweep slowly; keep the unneeded tools away; sweep precisely |
| DrillHole | precondition (precon) | point is clean; no hole exists; when human gives an order |
| | location (loc) | upper-right; center; bottom-right |
| | action sequence (act) | place→drill→left |
| | tool usages (tool) | driller, drilling arm, drilling machine |
| | human requirements (req) | slowly, keep away from unneeded tools, precisely |
| InstallScrew | precondition (precon) | a hole exists; no screw in hole. no unneccessary tool; hole size appropriate |
| | location (loc) | upper-right; center; bottom-right |
| | action sequence (act) | take screw→place→install→left |
| | tool usages (tool) | screwdriver; screw; install machine |
| | human requirements (req) | firmly; slowly; make surface clean |

(greater than 0.03) and precision (greater than 0.02) under the support of either the SVM algorithm or NB algorithm. Both the good performance and the relative advantages proved the effectiveness of exePlan in instruction disambiguation. In addition, SVM-based TC outperformed NB-based TC, proving SVM was more suitable than NB in understanding various task expressions. The disambiguation samples of task-related logic formulas are shown in Table 4. Based on keywords such as "remove", the potential sentences were located. Then based on the task-centered semantic analysis, the task-related sub-goals such as "CleanSpot" in sentence "remove the dust in middle with a brush" were identified due to the linguistic features such as {key word: remove, dust; pos:vb, nn; ...} and semantic features {MES action "remove"+ corresponding context "dust"}; while the irrelevant sentences such as "wait, I need

Table 3. Performance of the exePlan in instruction disambiguation

|  | SVM | | | | NB | | | |
|---|---|---|---|---|---|---|---|---|
|  | TC | | nonTC | | TC | | non-TC | |
|  | precision | recall | precision | recall | precision | recall | precision | recall |
| *CleanSpot* | 0.95 | 1.00 | 0.90 | 1.00 | 0.94 | 1.00 | 0.85 | 1.00 |
| *DrillHole* | 1.00 | 0.93 | 1.00 | 0.90 | 0.88 | 1.00 | 0.83 | 0.89 |
| *InstallScrew* | 0.98 | 0.98 | 0.97 | 0.98 | 1.00 | 0.85 | 0.92 | 0.76 |
| *tranCD* | 0.98 | 0.97 | 0.95 | 0.95 | 0.97 | 1.00 | 0.89 | 0.95 |
| *tranDC* | 1.00 | 0.93 | 1.00 | 0.88 | 0.97 | 1.00 | 0.88 | 0.95 |
| *tranCI* | 1.00 | 1.00 | 1.00 | 1.00 | 0.97 | 0.93 | 0.89 | 0.88 |
| *tranIC* | 1.00 | 1.00 | 0.94 | 1.00 | 0.97 | 1.00 | 0.89 | 0.88 |
| *tranDI* | 1.00 | 1.00 | 0.93 | 1.00 | 1.00 | 1.00 | 1.00 | 0.83 |
| *Average* | 0.99 | 0.98 | 0.96 | 0.96 | 0.96 | 0.97 | 0.89 | 0.89 |

TC: task-centered semantic analysis method. non-TC: non-task-centered semantic analysis method.

Table 4. Disambiguation samples of task-related logic formulas

|  | **Sub-goals** | **NOT Sub-goals** |
|---|---|---|
| *CleanSpot* | *Remove* the dust in middle with a brush. | Wait, I need to *remove* the tools on the surface. |
| *DrillHole* | At the bottom-right corner *drill* a hole. | Eh, the *drill* is missing. |
| *InstallScrew* | Then *install* a screw at the created hole. | First I will *install* a drill on your arm. |

to remove the tools on the surface" were filtered out due to the semantic features {incorrect task context: "tools", "tools+on surface"}.

*4.3. Evaluation of Intruction Interpretation and Executability Assessment*

Based on the extracted task-related logic formulas in instruction disambiguation, instruction interpretation was conducted by using exePlan to "translate" abstract NL instructions into machine-executable plans. Using these plans, the Baxter robot executed the instructor-assigned tasks. To evaluate the performance of the machine-execution-specification method included in exePlan, a baseline method was selected as *Literally Interpretation* method ("LI" for short), which was implemented in recent research of NL-based human-robot interactions [8][54]. In the LI method a plan was interpreted by literally understanding NL instructions, and the plan type was identified by strictly following human instructions in a first-order logic manner. For example, drill→clean denotes task type "clean". MES detection in this baseline method was keyword-based. Different from the LI method with literally task understanding, the exePlan method classified the plan type first and then selected the appropriate MES parameters for each logic formulas included in the identified plan. Each method's performance of instruction interpretation was assessed by identification precision/recall and MES/plan interpretation accuracy.

In plan identification, recall was calculated by the percentage of correctly-identified plans in all the volunteer-instructed plans. Precision was calculated by the percentage of correctly-identified specific-type plans in all the predicted specific-type plans. As Table 4 shows, exePlan's performance was good with an average precision of 0.93 and recall of 0.95. It outperformed the LI method by recall of 0.26 and precision of 0.28, proving the effectiveness of exePlan in identifying the task type.

In MES/plan interpretation evaluation, both the MES mapping accuracy and executable-plan proportion were calculated. MES mapping recall denoted the percentage of correctly-extracted MES in all the instructed MES and MES mapping precision denoted the percentage of correctly-extracted MES in all the extracted MES. Executable plan proportion denoted the percentage of the machine-executable plans after the plan interpretation. The executability threshold $\partial_0$ of the generated plan was 0.5, only greater than which a plan could be considered as machine-executable. As Table 5 shows, in both interpretation methods, MES mapping accuracy was good while the exePlan outperformed the LI method by precision of 0.3 and recall of 0.15, proving both the keyword-based method and the classification-based method could accurately find the available MES parameters. For the executable plan proportion, exePlan's performance was good and largely outperformed the LI method. The big difference in executable-plan generation could be explained by that usually in NL instructions due to the instructor's expression habits, some crucial MES parameters was missing. Even if all the mentioned MES parameters could be identified accurately, the generated plan was still machine-non-executable without instruction interpretation. The better performance in MES mapping and machine-executable plan generation proved that exePlan was effective in instruction interpretation.

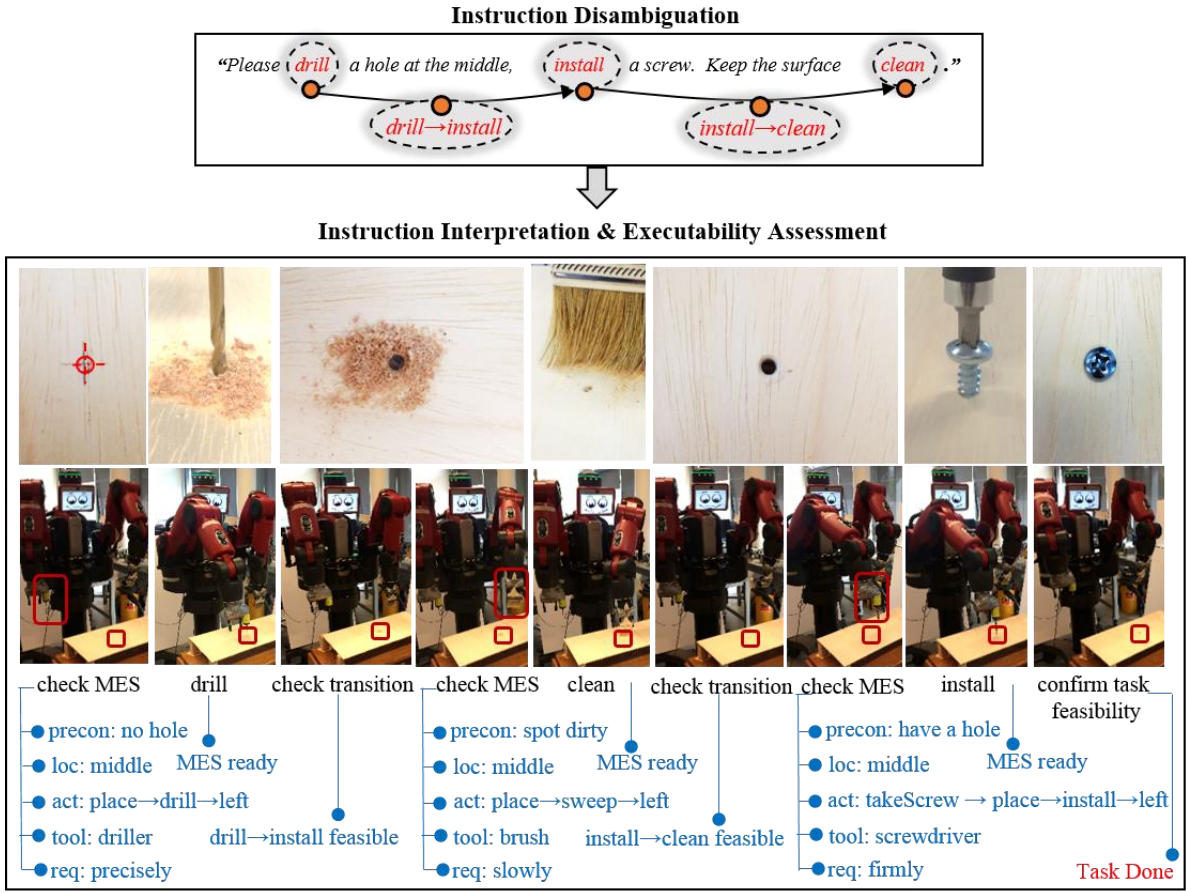

Fig. 10. Machine-executable plan generation. NL instructions, such as "please drill a hole at the middle, install a screw. Keep the surface clean" were recognized. With semantic analysis, sub-goals of the task were disambiguated as "drill, install, clean". Then the human-assigned task was interpreted as a machine-executable plan "drill →install →clean". Each sub-goal was specified with the execution parameters such as "action sequence, location, tool usage, precondition, human requirements". The feasibility of each sub-goal transition was assessed by checking the MES availability of the two sequential sub-goals.

The purpose of designing an executability assessment module is to detect non-executable plans and sub-goals. The detection accuracy depends on the threshold setting. Therefore, in the evaluation of the executability assessment module, only two general aspects were assessed: 1). the necessity of checking plan/sub-goal executability. 2). the effectiveness of checking plan/sub-goal executability. Based on the literal interpretation method that generated plans/sub-goals based on the information literally contained in the NL instructions, only 3% of the plans are executable. After being assessed by the executability module and launching the plan/sub-goal interpretation, the proportion of executable plans was increased to 95%. This significant increase proved that the module was necessary and effective in detecting the non-executable plans/sub-goals.

## 5. Conclusions & Future Work

In this paper, we have developed an exePlan method to generate machine-executable plans from ambiguous and abstract NL instructions. Two main problems have been solved. One is understanding of ambiguous instructions. To solve this problem, a task-centered semantic analysis method was developed. By understanding NL instructions given by 20 volunteers to an advanced Baxter robot, the effectiveness of semantic analysis of the exePlan method was validated. The second problem is interpretation of abstract NL instructions. To make the abstract task instruction executable by a machine, a machine-execution-specification method was developed to map the abstract NL instructions with the correct execution plan and the detailed MES parameters. By interpreting the abstract task instructions into machine-executable plans in the experiments, the effectiveness of this interpretation method was validated.

Table 4. Plan identification accuracy

|  | Literally Interpretation | | exePlan Interpretation | |
|---|---|---|---|---|
|  | precision | recall | precision | recall |
| Drill | 0.80 | 0.47 | 0.95 | 0.92 |
| Clean | 0.25 | 1.00 | 0.87 | 1.00 |
| Install | 0.96 | 0.55 | 0.96 | 0.92 |
| Average | 0.67 | 0.67 | 0.93 | 0.95 |

Table 5. Performance of MES/plan Interpretation

|  | Literally Interpretation | | | exePlan Interpretation | | |
|---|---|---|---|---|---|---|
|  | MES Mapping | | Executable-Plan Proportion | MES Mapping | | Executable-Plan Proportion |
|  | precision | recall |  | precision | recall |  |
| Drill | 0.94 | 0.75 | 0.00 | 0.95 | 1.00 | 1.00 |
| Clean | 1.00 | 0.88 | 0.00 | 1.00 | 0.93 | 0.92 |
| Install | 0.91 | 0.85 | 0.08 | 0.98 | 0.98 | 0.92 |
| Average | 0.95 | 0.83 | 0.03 | 0.98 | 0.97 | 0.95 |

In future work, focus will be placed on knowledge mapping from NL instruction to the real world by considering more complete constraints such as human/environmental factors, further making our exePlan practical.

## 6. Acknowledgements

We would like to thank Dr. Hao Zhang for providing the Baxter robot to help us with the experiments. We also would like to thank Mr. Jeremy Webb for assisting us with the experiment setting up.

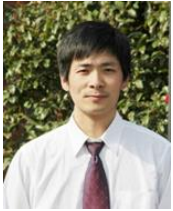
**Rui Liu** received the M.S. degree in active vibration control from Shanghai Jiao Tong University, Shanghai, China in 2013. Since January, 2014, he has been a Ph.D. student in Mechanical Engineering at Colorado School of Mines, Golden, Colorado, USA.

From June to December 2013, he was a Research Engineer at the Chinese Academy of Science, Chengdu, China. His current research interests include robot knowledge, computational linguistics and machine learning.
All my papers and PDFs could be found at:
Google Scholar: https://scholar.google.com/citations?user=X5X0IyQAAAAJ&hl=en&authuser=1
Research Gate: https://www.researchgate.net/profile/Rui_Liu99/contributions

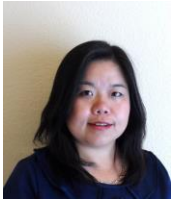
**Xiaoli Zhang** received the B.S. degree in Mechanical Engineering from Xi'an Jiaotong University, Xi'an, ShanXi, China in 2003, the M.S. degree in Mechatronics Engineering from Xi'an Jiaotong University in 2006, the Ph.D. degree in Biomedical Engineering from the University of Nebraska Lincoln, Lincoln, USA, in 2009.

Since 2013, she has been an Assistant Professor in the Mechanical Engineering Department, Colorado School of Mines, Golden, CO. She is the author of more than 50 articles, and 5 inventions. Her research interests include intelligent human-robot interaction, human intention awareness, robotics system design and control, haptics, and their applications in healthcare fields.